\documentclass[10pt,twocolumn,letterpaper]{article}

\usepackage{cvpr}      

\usepackage[accsupp]{axessibility}
\usepackage{graphicx}
\usepackage{amsmath}
\usepackage{amssymb}
\usepackage{booktabs}
\usepackage{multirow}
\usepackage{subcaption}
\usepackage{makecell}
\usepackage[table]{xcolor}

\usepackage[pagebackref,breaklinks,colorlinks]{hyperref}

\usepackage[capitalize]{cleveref}
\crefname{section}{Sec.}{Secs.}
\Crefname{section}{Section}{Sections}
\Crefname{table}{Table}{Tables}
\crefname{table}{Tab.}{Tabs.}

\def\cvprPaperID{496} 
\def\confName{CVPR}
\def\confYear{2023}

\begin{document}


\title{Iterative Geometry Encoding Volume for Stereo Matching}

\author{Gangwei Xu \quad Xianqi Wang \quad Xiaohuan Ding \quad Xin Yang\footnotemark[2]\\
{\normalsize School of EIC, Huazhong University of Science and Technology}\\
{\tt\small \{gwxu, xianqiw, dingxiaohuan, xinyang2014\}@hust.edu.cn}
}
\maketitle

\begin{abstract}
   Recurrent All-Pairs Field Transforms (RAFT) has shown great potentials in matching tasks. However, all-pairs correlations lack non-local geometry knowledge and have difficulties tackling local ambiguities in ill-posed regions. In this paper, we propose Iterative Geometry Encoding Volume (IGEV-Stereo), a new deep network architecture for stereo matching. The proposed IGEV-Stereo builds a combined geometry encoding volume that encodes geometry and context information as well as local matching details, and iteratively indexes it to update the disparity map. To speed up the convergence, we exploit GEV to regress an accurate starting point for ConvGRUs iterations. Our IGEV-Stereo ranks $1^{st}$ on KITTI 2015 and 2012 (Reflective) among all published methods and is the fastest among the top 10 methods. In addition, IGEV-Stereo has strong cross-dataset generalization as well as high inference efficiency. We also extend our IGEV to multi-view stereo (MVS), i.e. IGEV-MVS, which achieves competitive accuracy on DTU benchmark.
   Code is available at \textcolor{magenta}{https://github.com/gangweiX/IGEV}.
\end{abstract}

{
\renewcommand{\thefootnote}{\fnsymbol{footnote}}
\footnotetext[2]{Corresponding author.}}

\section{Introduction}
\label{sec:intro}
Inferring 3D scene geometry from captured images is a fundamental task in computer vision and graphics with applications ranging from 3D reconstruction, robotics
and autonomous driving. Stereo matching which aims to reconstruct dense 3D representations from two images with calibrated cameras is a key technique for reconstructing 3D scene geometry.

Many learning-based stereo methods\cite{psmnet, gwcnet,raft-stereo, acvnet, fast-acv} have been proposed in the literature. The popular representative is PSMNet\cite{psmnet} which apply a 3D convolutional encoder-decoder to aggregate and regularize a 4D cost volume and then use $soft \; argmin$ to regress the disparity map from the regularized cost volume. Such 4D cost volume filtering-based methods can effectively explore stereo geometry information and achieve impressive performance on several benchmarks. However, it usually demands a large amount of 3D convolutions for cost aggregation and regularization, and in turn yield high computational and memory cost. As a result, it can hardly be applied to high-resolution images and/or large-scale scenes. 

\begin{figure}[t]
\centering
{\includegraphics[width=1.0\linewidth]{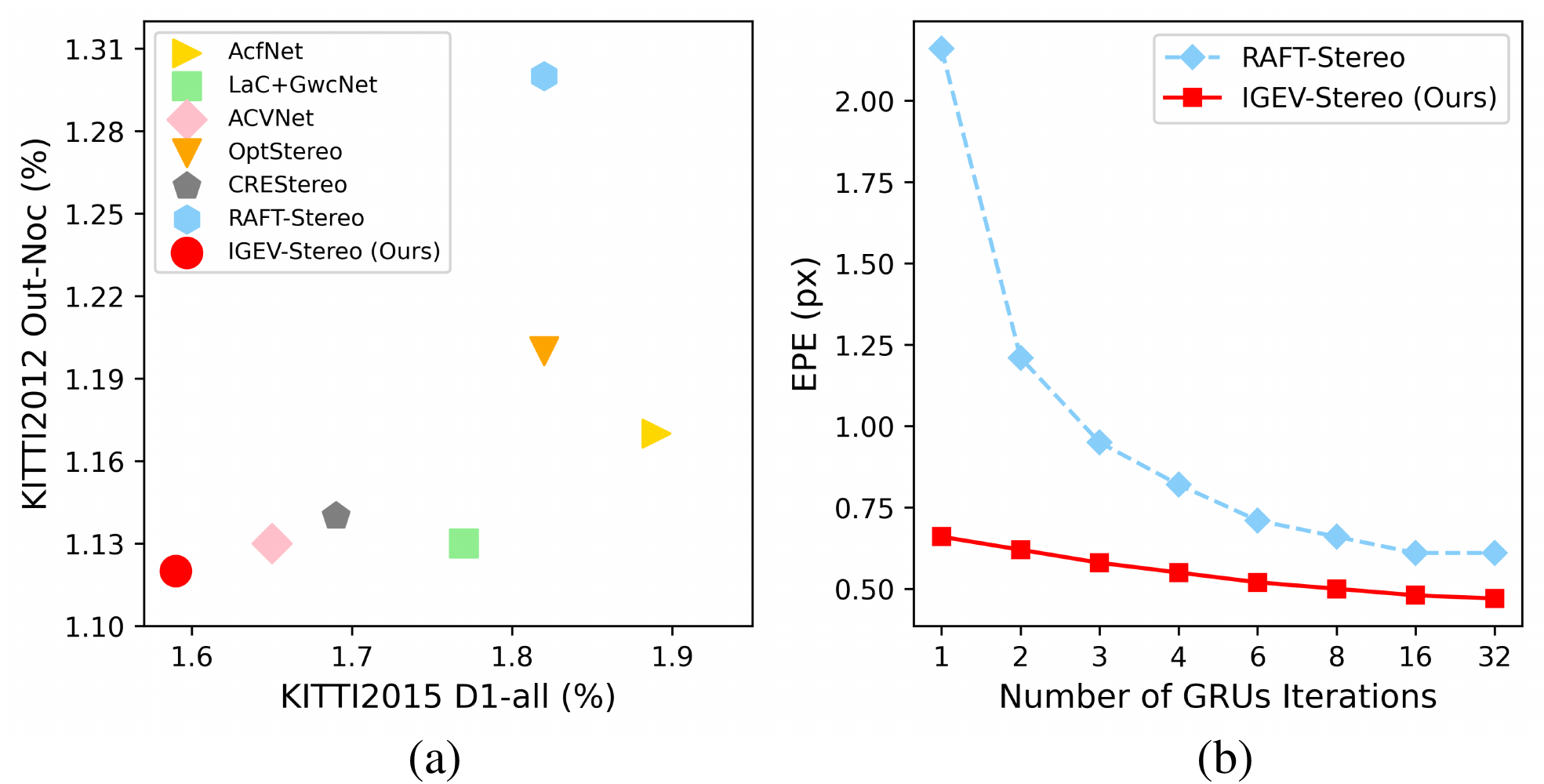}}
\caption{(a) Comparison with state-of-the-art stereo methods\cite{crestereo,acvnet,leastereo,optstereo,lsp,acfnet} on KITTI 2012 and 2015 leaderboards. 
(b) Performance comparison with RAFT-Stereo\cite{raft-stereo} on Scene Flow test set as the number of iterations changes.}\label{fig:ranking}
\vspace{-15pt}
\end{figure}

\begin{figure*}
\centering
\includegraphics[width=0.95\textwidth]{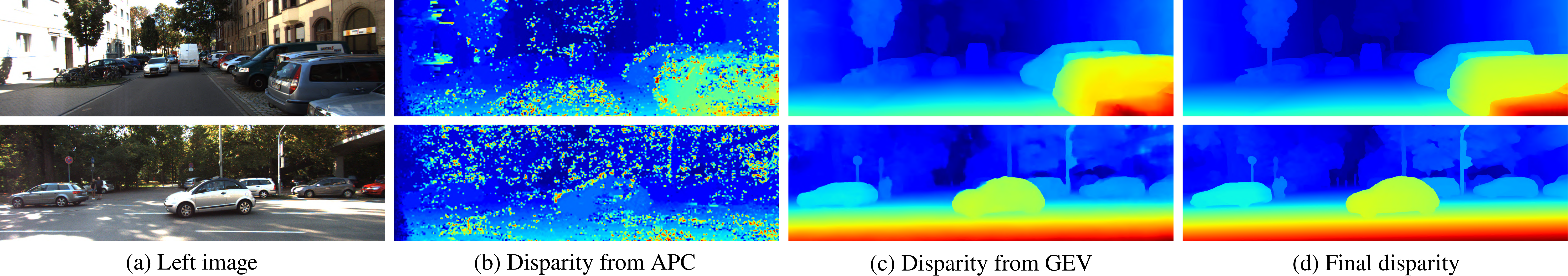} 
\caption{(a) Input images from KITTI 2015. Illustration of (b) disparity regressed from All-pairs Correlations (APC) in RAFT-Stereo\cite{raft-stereo}, (c) disparity regressed from our Geometry Encoding Volume (GEV), (d) our final disparity. The APC lacks non-local geometry knowledge and thus has difficulties tackling local ambiguities in ill-posed region. 
We take full advantage of cost filtering and iterative optimization: 1) exploiting 3D CNN to filter cost volume and obtain the strong scene representation and the initial disparity with smooth edges, 2) exploiting ConvGRUs to optimize the initial disparity to recover object edges and details.}
\label{fig:apc_gev}
\vspace{-10pt}
\end{figure*}

Recently, iterative optimization-based methods\cite{raft, raft-stereo, crestereo, cer-mvs, optstereo} have exhibited attractive performance on both high resolution images and standard benchmarks. Different from existing methods, iterative methods bypass the computationally expensive cost aggregation operations and progressively update the disparity map by repeatedly fetching information from a high-resolution 4D cost volume. Such solution enables the direct usage of high-resolution cost volume and hence is applicable to high-resolution images. For instance, RAFT-Stereo\cite{raft-stereo} exploits a multi-level Convolutional Gated Recurrent Units (ConvGRUs)\cite{gru} to recurrently update the disparity field using local cost values retrieved from all-pairs correlations (APC). 

However, without cost aggregation the original cost volume lacks non-local geometry and context information (see Fig.~\ref{fig:apc_gev} (b)). As a result, existing iterative methods have difficulties tackling local ambiguities in ill-posed regions, such as occlusions, texture-less regions and repetitive structures. Even though, the ConvGRU-based updater can improve the predicted disparities by incorporating context and geometry information from context features and hidden layers, such limitation in the original cost volume greatly limits the effectiveness of each iteration and in turn yield a large amount of ConvGRUs iterations for satisfactory performance. 

\begin{figure*}[t]
    \centering
    \includegraphics[width=0.9\linewidth]{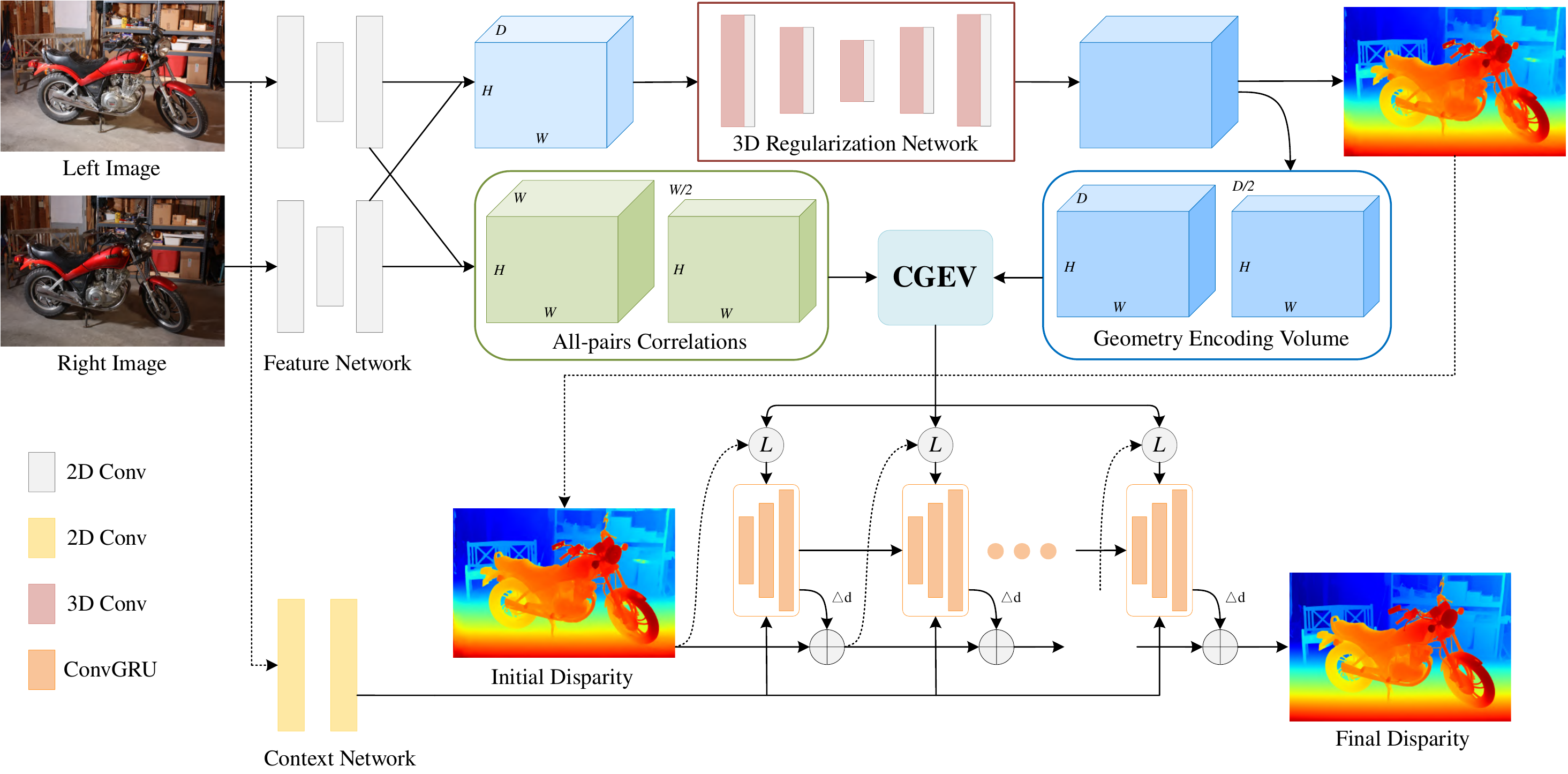}
    \caption{Overview of our proposed IGEV-Stereo. The IGEV-Stereo first builds a Geometry Encoding Volume (GEV) which encodes geometry and context information through 3D CNN, and combines it with All-pairs Correlations (APC) to form a Combined Geometry Encoding Volume (CGEV). Then we regress an initial disparity from GEV and iteratively update it using local cost values retrieved from CGEV through ConvGRUs.}
    \label{fig:igev_stereo}
    \vspace{-10pt}
\end{figure*}

We claim that cost filtering-based methods and iterative optimization-based methods have complementary advantages and limitations. The former can encode sufficient non-local geometry and context information in the cost volume which is essential for  disparity prediction in particular in challenging regions. The latter can avoid high computational and memory cost for 3D cost aggregation, yet are less capable in ill-posed regions based only on all-pairs correlations. To combine complementary advantages of the two methods, we propose Iterative Geometry Encoding Volume (IGEV-Stereo), a new paradigm for stereo matching (see Fig. \ref{fig:igev_stereo}).
To address ambiguities caused by ill-posed regions, we compute a Geometry Encoding Volume (GEV) by aggregating and regularizing a cost volume using an extremely lightweight 3D regularization network. Compared to all-pairs correlations of RAFT-Stereo\cite{raft-stereo}, our GEV encodes more geometry and context of the scene after aggregation, shown in Fig.~\ref{fig:apc_gev} (c). A potential problem of GEV is that it could suffer from over-smoothing at boundaries and tiny details due to the 3D regularization network. To complement local correlations, we combine the GEV and all-pairs correlations to form a Combined Geometry Encoding Volume (CGEV) and input the CGEV into the ConvGRU-based update operator for iterative disparity map optimization. 

Our IGEV-Stereo outperforms RAFT-Stereo in terms of both accuracy and efficiency. The performance gains come from two aspects. First, our CGEV provides more comprehensive yet concise information for ConvGRUs to update, yielding more effective optimization in each iteration and in turn could significantly reduce the amount of ConvGRUs iterations. As shown in Fig.~\ref{fig:ranking}, our method achieves even smaller EPE (i.e., 0.58) using only 3 ConvGRUs iterations (i.e.,100ms totally for inference) than RAFT-Stereo using 32 ConvGRUs iterations (i.e., EPE of 0.61 and 440ms for inference). Second, our method regresses an initial disparity map from the GEV via $soft \; argmin$ which could provide an accurate starting point for the ConvGRU-based update operator, and in turn yield a fast convergence. In comparison,  RAFT-Stereo starts disparity prediction from an initial starting point $\mathbf{d}_0$=0, which demands a large number ConvGRUs iterations to achieve an optimized result.

We demonstrate the efficiency and effectiveness of our method on several stereo benchmarks. Our IGEV-Stereo achieves the state-of-the-art EPE of 0.47 on Scene Flow\cite{dispNetC} and ranks $1^{st}$ on KITTI 2015\cite{kitti2015} and 2012 (Reflective)\cite{kitti2012} leaderboards among all the published methods. Regarding the inference speed, our IGEV-Stereo is the fastest among the top 10 methods on KITTI leaderboards. IGEV-Stereo also exhibits better cross-dataset generalization ability than most existing stereo networks. When trained only on synthetic data Scene Flow, our IGEV-Stereo performs very well on real datasets Middlebury\cite{middlebury} and ETH3D\cite{eth3d}. We also extend our IGEV to MVS, i.e. IGEV-MVS, which achieves competitive accuracy on DTU\cite{dtu}.


\section{Related Work}
\label{sec:related_work}
\noindent\textbf{Cost Filtering-based Methods} To improve the representative ability of a cost volume, most existing learning-based stereo methods\cite{psmnet,nie2019multi, multilevel, segstereo, sspcv,  bgnet, deeppruner, decomposition} construct a cost volume using powerful CNN features. However, the cost volume could still suffer from the ambiguity problem in occluded regions, large texture-less/reflective regions and repetitive structures. The 3D convolutional networks have exhibited great potential in regularizing or filtering the cost volume, which can propagate reilable sparse matches to ambiguous and noisy regions. GCNet\cite{gcnet} firstly uses 3D encoder-decoder architecture to regularize a 4D concatenation volume. PSMNet\cite{psmnet} proposes a stacked hourglass 3D CNN in conjunction with intermediate supervision to regularize the concatenation volume. GwcNet\cite{gwcnet} and ACVNet\cite{acvnet} propose the group-wise correlation volume and the attention concatenation volume, respectively, to improve the expressiveness of the cost volume and in turn improve the performance in ambiguous regions. GANet\cite{ganet} designs a semi-global aggregation layer and a local guided aggregation layer to further improve the accuracy. However, the high computational and memory cost of 3D CNNs often prevent these models from being applied to high-resolution cost volumes. To improve efficiency, several cascade methods\cite{cfnet, cascade, fast-acv} have been proposed. CFNet\cite{cfnet} and CasStereo\cite{cascade} build a cost volume pyramid in a coarse-to-fine manner to progressively narrow down the predicted disparity range. Despite their impressive performance, the coarse-to-fine method inevitably involves accumulated errors at coarse resolutions. 

\noindent\textbf{Iterative Optimization-based Methods} Recently, many iterative methods\cite{raft,raft-stereo,itermvs} have been proposed and achieved impressive performance in matching tasks. RAFT-Stereo\cite{raft-stereo} proposes to recurrently update the disparity field using local cost values retrieved from the all-pairs correlations. However, the all-pair correlations lack non-local information and have difficulties in tackling local ambiguities in ill-posed regions.  Our IGEV-Stereo also adopts ConvGRUs as RAFT-Stereo\cite{raft-stereo} to iteratively update the disparities. Different from RAFT-Stereo\cite{raft-stereo}, we construct a CGEV which encodes non-local geometry and context information, and local matching details to significantly improve the effectiveness of each ConvGRUs iteration. In addition, we provide a better initial disparity map for the ConvGRUs updater to start, yielding a much faster convergence than RAFT-Stereo\cite{raft-stereo}.   


\section{Method}
\label{sec:method}
In this section, we detail the structure of IGEV-Stereo (Fig.~\ref{fig:igev_stereo}), which consists of a multi-scale feature extractor, a combined geometry encoding volume, a ConvGRU-based update operator and a spatial upsampling module.

\subsection{Feature Extractor}

Feature extractor contains two parts: 1) a feature network which extracts multi-scale features for cost volume construction and guiding cost aggregation, and 2) a context network which extracts multi-scale context features for ConvGRUs hidden state initialization and updating.

\textbf{Feature Network.} Given the left and the right images $\mathbf{I}_{l(r)}\in\mathbb{R}^{{3}\times{H}\times{W}}$, we first apply the MobileNetV2 pretrained on ImageNet\cite{imagenet} to scale $\mathbf{I}_{l(r)}$ down to 1/32 of the original size, then use upsampling blocks with skip-connections to recover them up to 1/4 scale, resulting in multi-scale features \{$\mathbf{f}_{l,i}(\mathbf{f}_{r,i})\in\mathbb{R}^{{C_i}\times{\frac{H}{i}}\times{\frac{W}{i}}}$\} ($i$=4, 8, 16, 32 and $C_i$ for feature channels). The $\mathbf{f}_{l,4}$ and $\mathbf{f}_{r,4}$ are used to construct the cost volume. And the $\mathbf{f}_{l,i}$ ($i$=4, 8, 16, 32) are used as guidance for 3D regularization network.

\textbf{Context Network.} Following RAFT-Stereo\cite{raft-stereo}, the context network consists of a series of residual blocks and downsampling layers, producing multi-scale context features at 1/4, 1/8 and 1/16 of the input image resolution with 128 channels. The multi-scale context features are used to initialize the hidden state of the ConvGRU-based update operator and also inserted into the ConvGRUs at each iteration.

\subsection{Combined Geometry Encoding Volume}
Given the left features $\mathbf{f}_{l,4}$ and right features $\mathbf{f}_{r,4}$ extracted from $\mathbf{I}_{l}$ and $\mathbf{I}_{r}$, we construct a group-wise correlation volume\cite{gwcnet} that splits features $\mathbf{f}_{l,4}$ ($\mathbf{f}_{r,4}$) into $N_g$ ($N_g$=8) groups along the channel dimension and compute correlation maps group by group,
\begin{equation}
\mathbf{C}_{corr}(g,d,x,y)=\frac{1}{N_c/N_g}\langle\mathbf{f}_{l,4}^g(x,y), \mathbf{f}_{r,4}^g(x-d,y)\rangle, 
\end{equation}
where $\langle\cdot,\cdot\rangle$ is the inner product, $d$ is the disparity index, $N_c$ denotes the number of feature channels.
A cost volume $\mathbf{C}_{corr}$ based on only feature correlations lacks the ability to capture global geometric structure. To address this problem, we further process $\mathbf{C}_{corr}$ using a lightweight 3D regularization network $\mathbf{R}$ to obtain the geometry encoding volume $\mathbf{C}_{G}$ as,
\begin{equation}
\mathbf{C}_{G}=\mathbf{R}(\mathbf{C}_{corr})
\end{equation}
The 3D regularization network $\mathbf{R}$ is based on a lightweight 3D UNet that consists of three down-sampling blocks and three up-sampling blocks. Each down-sampling block consists of two $3\times3\times\!3$ 3D convolution. The number of channels of the three down-sampling blocks are 16, 32, 48 respectively. Each up-sampling block consists of a $4\times4\times4$ 3D transposed convolution and two $3\times3\times3$ 3D convolutions.
We follow CoEx\cite{coex}, which excites the cost volume channels with weights computed from the left features for cost aggregation. For a $\frac{D}{i}\times\frac{H}{i}\times\frac{W}{i}$ cost volume $\mathbf{C}_i$ ($i$=4, 8, 16 and 32) in cost aggregation, the guided cost volume excitation is expressed as,
\begin{equation}
\mathbf{C}_{i}^{'}=\sigma(\mathbf{f}_{l,i})\odot\mathbf{C}_i ,
\end{equation}
where $\sigma$ is the sigmoid function, $\odot$ denotes the Hadamard Product. The 3D regularization network, which inserts guided cost volume excitation operation, can effectively infer and propagate scene geometry information, leading to a geometry encoding volume. We also calculate all-pairs correlations between corresponding left and right features to obtain local feature correlations.

To increase the receptive field, we pool the disparity dimension using 1D average pooling with a kernel size of 2 and a stride of 2 to form a two-level $\mathbf{C}_{G}$ pyramid and all-pairs correlation volume $\mathbf{C}_{A}$ pyramid. Then we combine the $\mathbf{C}_{G}$ pyramid and $\mathbf{C}_{A}$ pyramid to form a combined geometry encoding volume.


\subsection{ConvGRU-based Update Operator}
We apply $soft \; argmin$ to regress an initial starting disparity $\mathbf{d}_0$ from the geometry encoding volume $\mathbf{C}_{G}$ according to Equ.~\ref{equ:soft},
\begin{equation}
\begin{aligned}
\mathbf{d}_{0} = \sum\limits_{d=0}^{D-1} d \times Softmax(\mathbf{C}_{G}(d)),
\end{aligned}
\label{equ:soft}
\end{equation}
where $d$ is a predetermined set of disparity indices at 1/4 resolution. Then from $\mathbf{d}_{0}$, we use three levels of ConvGRUs to iteratively update the disparity (shown in Fig. \ref{fig:igev_stereo}). This setup facilitates a fast convergence of iterative disparity optimization. The hidden state of three levels of ConvGRUs are initialized from the multi-scale context features.

For each iteration, we use the current disparity $\mathbf{d}_k$ to index from the combined geometry encoding volume via linear interpolation, producing a set of geometry features $\mathbf{G}_f$.
The $\mathbf{G}_f$ is computed by,
\begin{equation} 
\begin{aligned}
\mathbf{G}_{f}=\sum\limits_{i=-r}^{r}\text{Concat}\{\mathbf{C}_{G}(\mathbf{d}_k\!+\!i),\mathbf{C}_{A}(\mathbf{d}_k\!+\!i),\\\mathbf{C}_{G}^{p}({\mathbf{d}_k}/2\!+\!i), \mathbf{C}_{A}^{p}({\mathbf{d}_k}/2\!+\!i)\},
\end{aligned}
\end{equation}
where $\mathbf{d}_k$ is the current disparity, $r$ is indexing radius, and $p$ denotes the pooling operation. These geometry features and current disparity prediction $\mathbf{d}_k$ are passed through two encoder layers and then concatenated with $\mathbf{d}_k$ to form $x_k$. Then we use ConvGRUs to update the hidden state $h_{k-1}$ as RAFT-Stereo\cite{raft-stereo},
\begin{equation}
\begin{aligned}
x_k = & \; [\text{Encoder}_{g}(\mathbf{G}_f), \text{Encoder}_{d}(\mathbf{d}_k),\mathbf{d}_k]\\
z_k = & \;\sigma(\text{Conv}([h_{k-1}, x_k], W_z) + c_k), \\
r_k = & \;\sigma(\text{Conv}([h_{k-1}, x_k], W_r) + c_r), \\
\Tilde{h}_k = & \,\tanh(\text{Conv}([r_k \odot h_{k-1}, x_k], W_h) + c_h), \\
h_k = & \;(1-z_k) \odot h_{k-1} + z_k \odot \Tilde{h}_k,
\end{aligned}
\end{equation}
where $c_k$, $c_r$, $c_h$ are context features generated from the context network. The number of channels in the hidden states of ConvGRUs is 128, and the number of channels of the context feature is also 128. The $\text{Encoder}_{g}$ and $\text{Encoder}_{d}$ consist of two convolutional layers respectively. Based on the hidden state $h_k$, we decode a residual disparity $\triangle \mathbf{d}_k$ through two convolutional layers, then we update the current disparity,
\begin{equation}
\begin{aligned}
\mathbf{d}_{k+1} = & \; \mathbf{d}_{k} + \triangle \mathbf{d}_k
\end{aligned}
\end{equation}
\subsection{Spatial Upsampling}
We output a full resolution disparity map by the weighted combination of the predicted disparity $\mathbf{d}_k$ at 1/4 resolution. Different from RAFT-Stereo\cite{raft-stereo} which predicts weights from the hidden state ${h_k}$ at 1/4 resolution, we utilize the higher resolution context features to obtain the weights. We convolve the hidden state to generate features and then upsample them to 1/2 resolution. The upsampled features are concatenated with $\mathbf{f}_{l,2}$ from left image to produce weights $\mathbf{W}\in\mathbb{R}^{H\times{W}\times9}$. We output the full resolution disparities by the weighted combination of their coarse resolution neighbors.

\subsection{Loss Function}
We calculate the smooth L1 loss\cite{psmnet} on initial disparity $\mathbf{d}_0$ regressed from GEV:
\begin{equation}
    \mathcal{L}_{init} = Smooth_{L_1}(\mathbf{d}_0-\mathbf{d}_{gt})
\end{equation}
where $\mathbf{d}_{gt}$ represents the ground truth disparity. We calculate the L1 loss on all predicted disparities $\{\mathbf{d}_{i}\}_{i=1}^{N}$.
We follow \cite{raft-stereo} to exponentially increase the weights, and the total loss is defined as:
\begin{equation}
    \mathcal{L}_{stereo} = \mathcal{L}_{init} + \sum_{i=1}^{N} \gamma^{N-i} ||\mathbf{d}_i-\mathbf{d}_{gt}||_1,
\end{equation}
where $\gamma=0.9$, and ${\mathbf{d}}_{gt}$ represent ground truth. 

\begin{table*}
  \centering
  \begin{tabular}{l|cccc|cc|cc}
    \toprule
    Model &\makecell{{All-pairs} \\ {correlations}} &{GEV} &\makecell{{Init disp} \\ {from GEV}} &\makecell{{Supervise} \\ {for GEV}} & \makecell{EPE \\(px)} &\makecell{\textgreater3px\\ (\%)}  & \makecell{Time \\(s)} &\makecell{Params. \\(M)}\\
    \midrule
    \midrule
    Baseline & \checkmark & &  & &0.56 &2.85  &0.36  &12.02 \\
    \midrule
    G & & \checkmark & & &0.51 &2.68  &0.37 &12.60 \\
    G+I & &\checkmark &\checkmark & &0.50 &2.62  &0.37  &12.60 \\
    G+I+S & &\checkmark  &\checkmark &\checkmark &0.48 &2.51  &0.37  &12.60\\
    Full model (IGEV-Stereo) &\checkmark &\checkmark &\checkmark &\checkmark &\cellcolor{yellow!25}\textbf{0.47} &\cellcolor{yellow!25}\textbf{2.47} &0.37  &12.60\\
    \bottomrule
  \end{tabular}
  \vspace{-2mm}
  \caption{
  Ablation study of proposed networks on the Scene Flow test set. GEV denotes Geometry Encoding Volume. The baseline is RAFT-Stereo using MobileNetV2 100 as backbone. The time is the inference time for 960$\times$540 inputs.}
  \label{tab:effectiveness}
  \vspace{-5pt}
\end{table*}

\section{Experiment}
\textbf{Scene Flow}\cite{dispNetC} is a synthetic dataset containing 35, 454 training pairs and 4,370 testing pairs with dense disparity maps. We use the Finalpass of Scene Flow, since it is more like real-world images than the Cleanpass, which contains more motion blur and defocus. 

\textbf{KITTI 2012}\cite{kitti2012} and \textbf{KITTI 2015}\cite{kitti2015} are datasets for real-world driving scenes. KITTI 2012 contains 194 training pairs and 195 testing pairs, and KITTI 2015 contains 200 training pairs and 200 testing pairs. Both datasets provide sparse ground-truth disparities obtained with LIDAR.

\textbf{Middlebury 2014}\cite{middlebury} is an indoor dataset, which provides 15 training pairs and 15 testing pairs, where some samples are under inconsistent illumination or color conditions. All of the images are available in three different resolutions.
\textbf{ETH3D}\cite{eth3d} is a gray-scale dataset with 27 training pairs and 20 testing pairs. We use the training pairs of Middlebury 2014 and ETH3D to evaluate cross-domain generalization performance.



\subsection{Implementation Details}
We implement our IGEV-Stereo with PyTorch and perform our experiments using NVIDIA RTX 3090 GPUs. For all training, we use the AdamW\cite{adamw} optimizer and clip gradients to the range [-1, 1]. On Scene Flow, we train IGEV-Stereo for 200k steps with a batch size of 8. On KITTI, we finetune the pre-trained Scene Flow model on the mixed KITTI 2012 and KITTI 2015 training image pairs for 50k steps. We randomly crop images to $320\times736$ and use the same data augmentation as \cite{raft-stereo} for training. The indexing radius is set to 4. For all experiments, we use a one-cycle learning rate schedule with a learning rate of 0.0002, and we use 22 update iterations during training.

\begin{table} \small
  \centering
  \begin{tabular}{lcccccc}
    \toprule
    \multirow{2}{*}{Model} &\multicolumn{6}{c}{Number of Iterations}\\
    \cline{2-7}
    & 1 & 2 & 3 & 4 & 8 & 32 \\
    \midrule
    RAFT-Stereo\cite{raft-stereo} &2.16 &1.21 &0.95 &0.82 &0.66  &0.61 \\
    G  &0.98 &0.73 &0.66 &0.62 &0.54 &0.51\\
    G+I+S  &0.67 &0.63 &0.59 &0.56 &0.51 &0.48\\
    Full model  &\cellcolor{yellow!25}\textbf{0.66} &\cellcolor{yellow!25}\textbf{0.62} &\cellcolor{yellow!25}\textbf{0.58} &\cellcolor{yellow!25}\textbf{0.55} &\cellcolor{yellow!25}\textbf{0.50} &\cellcolor{yellow!25}\textbf{0.47}\\
    \bottomrule
  \end{tabular}
  \vspace{-2mm}
  \caption{
  Ablation study for number of iterations.}
  \label{tab:iter}
  \vspace{-5pt}
\end{table}

\begin{table} \footnotesize
  \centering
  \begin{tabular}{lccc}
    \toprule
    Experiment &Variations &Scene Flow & Params.(M)
    \\
     \midrule
    \multirow{2}{*}{GEV Reso.} & 1/8 & 0.49  & 12.71 \\
     & \underbar{1/4} &0.47 &12.60 \\
     \midrule
     \multirow{3}{*}{Backbone} & \underbar{MobileNetV2 100} &0.47  &12.60 \\
     & MobileNetV2 120d &0.46 &15.05 \\
     & ConvNeXt-B &0.45 &18.04 \\
    \bottomrule
  \end{tabular}
  \vspace{-2mm}
  \caption{
  Ablation experiments. Settings used in our final model are underlined.}
  \label{tab:ablation}
  \vspace{-15pt}
\end{table}

\begin{table*} \small
    \centering
    \begin{tabular}{lcccccccc}
     \toprule
     Method &PSMNet\cite{psmnet} & GwcNet\cite{gwcnet}  &GANet\cite{ganet} &CSPN\cite{cspn}  &LEAStereo\cite{leastereo} &ACVNet\cite{acvnet} & IGEV-Stereo (Ours)\\
     EPE (px) &1.09 &0.76 &0.84 &0.78 &0.78 &0.48 &\cellcolor{yellow!25}\textbf{0.47}\\
    \bottomrule
    \end{tabular}
    \vspace{-2mm}
    \caption{Quantitative evaluation on Scene Flow test set. \textbf{Bold}: Best.}
\label{tab:sceneflow}
\end{table*}

\begin{table*}
    \centering
    \begin{tabular}{l|cccccc|ccc|c}
     \toprule
     \multirow{2}{*}{Method} & \multicolumn{6}{c|}{KITTI 2012\cite{kitti2012}} & \multicolumn{3}{c|}{ KITTI 2015\cite{kitti2015}} & \multirow{2}{*}{\makecell{Run-time\\(s)}} \\
     & 2-noc & 2-all & 3-noc & 3-all & \thead{EPE \\ noc} & \thead{EPE\\all} & D1-bg & D1-fg & D1-all & \\
     \midrule

    PSMNet\cite{psmnet}  & {2.44} & {3.01} &1.49 & 1.89 & 0.5 & 0.6 & 1.86 & 4.62 & 2.32 & 0.41 \\
    GwcNet\cite{gwcnet} & 2.16 & 2.71 & 1.32 & 1.70 & 0.5 &0.5 & 1.74 & 3.93 & 2.11 & 0.32 \\
    GANet-deep\cite{ganet}  & 1.89 & 2.50 & 1.19 & 1.60 & 0.4 &0.5 & 1.48 & 3.46 & 1.81 & 1.80 \\
    AcfNet\cite{acfnet} & 1.83 & 2.35 &{1.17}  & {1.54} &  {0.5} & {0.5} & {1.51}  & {3.80} & {1.89} & 0.48\\
    HITNet\cite{hitnet} & {2.00} & {2.65} &{1.41}  & {1.89} &  0.4 & {0.5} & {1.74}  & {3.20} & {1.98} & 0.02\\
    EdgeStereo-V2\cite{edgestereo} &2.32 &2.88 &1.46 &1.83 & 0.4 & 0.5 &1.84 &3.30 & 2.08 & 0.32 \\
    CSPN\cite{cfnet} & {1.79} & {2.27} &{1.19}  & {1.53} &- &- & {1.51}  & {2.88} & {1.74} & 1.00\\
    LEAStereo\cite{leastereo}  & 1.90 & 2.39 & 1.13 & 1.45 & 0.5 &0.5 & 1.40 & 2.91 & 1.65 & 0.30 \\
    ACVNet\cite{acvnet} & 1.83 & 2.35 & 1.13 & 1.47 & 0.4 &0.5 & \cellcolor{yellow!25}\textbf{1.37} & 3.07 & 1.65 & 0.20 \\
    CREStereo\cite{crestereo} &1.72 &2.18 &1.14 & 1.46 & 0.4 & 0.5 &1.45 &2.86 &1.69 &0.41 \\
    RAFT-Stereo\cite{raft-stereo} &1.92 &2.42 &1.30 & 1.66 & 0.4 & 0.5 &1.58 &3.05 &1.82 &0.38 \\
    IGEV-Stereo (Ours) &\cellcolor{yellow!25} \textbf{1.71} & \cellcolor{yellow!25}\textbf{2.17} &\cellcolor{yellow!25}\textbf{1.12} & \cellcolor{yellow!25}\textbf{1.44} & 0.4 &\cellcolor{yellow!25} \textbf{0.4} &{1.38} &\cellcolor{yellow!25}\textbf{2.67} & \cellcolor{yellow!25}\textbf{1.59} &0.18 \\
    \bottomrule
    \end{tabular}
    \vspace{-2mm}
    \caption{Quantitative evaluation on KITTI 2012~\cite{kitti2012} and KITTI 2015~\cite{kitti2015}. The IGEV-Stereo runs 16 updates at inference. \textbf{Bold}: Best.}
\label{tab:kitti}
\vspace{-10pt}
\end{table*}

\begin{table} \small
  \centering
  \begin{tabular}{c|c|c|c|c|c}
    \toprule
    Method & Iters & 2-noc & 2-all &3-noc &3-all \\
    \midrule
    \multirow{3}{*}{RAFT-Stereo\cite{raft-stereo}} & 8 & 9.98 & 11.95 & 6.64 & 8.04 \\
    & 16 & 8.83 & 10.35 & 5.76 & 6.84 \\
    & 32 & 8.41 & 9.87 & 5.40 & 6.48 \\
    \midrule
    \multirow{3}{*}{IGEV-Stereo} & 8 & 8.30 & 9.82 &4.88 & 5.87 \\
    & 16 & 7.57 & 8.80 & 4.35 & 5.00 \\
    & 32 & 7.29 & 8.48 & 4.11 & 4.76\\
    \bottomrule
  \end{tabular}
  \vspace{-2mm}
  \caption{Evaluation in the reflective regions (ill-posed regions) of KITTI 2012 benchmark. Iters denotes iteration number.} 
  \label{tab:reflective}
  \vspace{-10pt}
\end{table}

\subsection{Ablation Study}

\begin{figure*}
\centering
\includegraphics[width=0.95\textwidth]{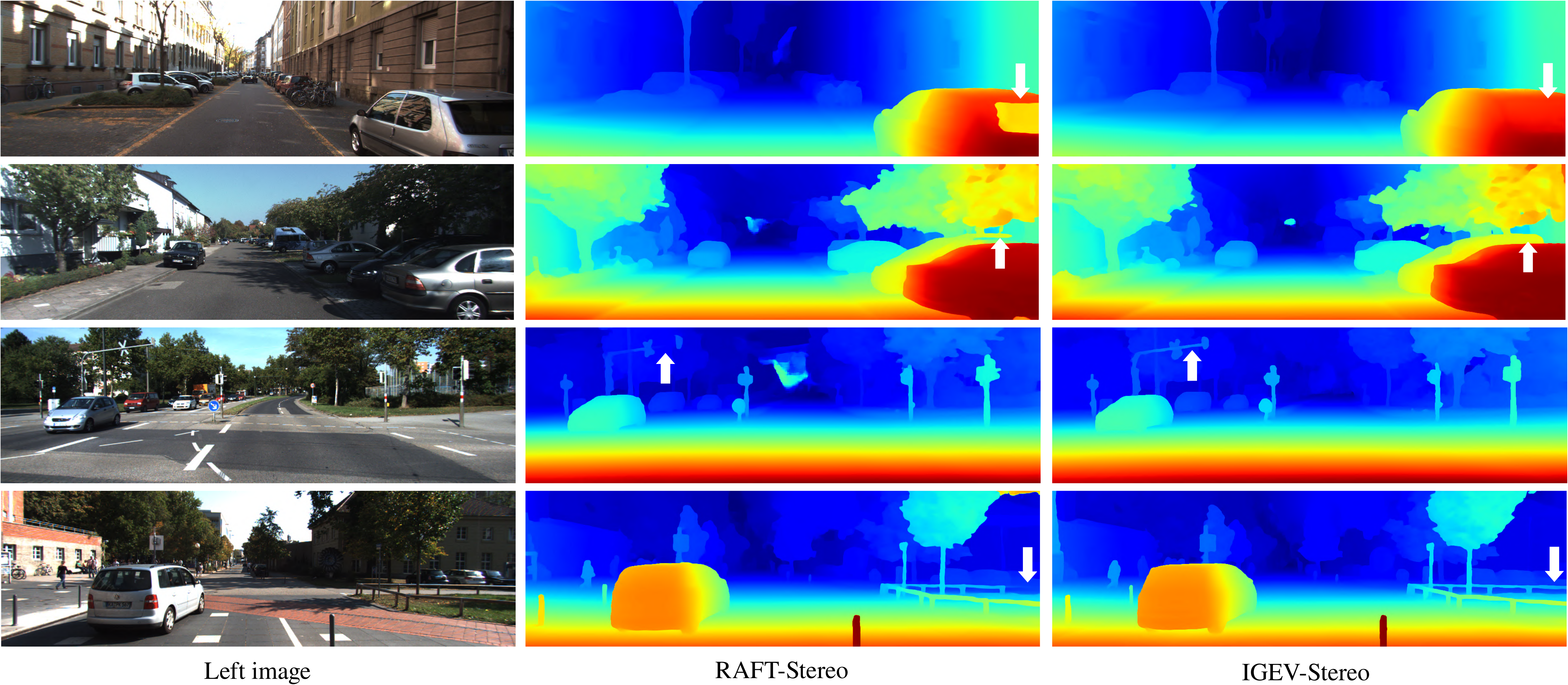} 
\caption{Qualitative results on the test set of KITTI. The first two columns show results on KITTI 2012, and the last two columns show results on KITTI 2015. Our IGEV-Stereo performs very well in textureless and detailed regions.}
\label{fig:kitti}
\vspace{-10pt}
\end{figure*}

\textbf{Effectiveness of CGEV.} We explore the best settings for the combined geometry encoding volume (CGEV) and exam its effectiveness. For all models in these experiments, we perform 32 iterations of ConvGRUs updating at inference. We take RAFT-Stereo as our baseline, and replace its original backbone with MobileNetV2 100. As shown in Tab.~\ref{tab:effectiveness}, the proposed GEV can significantly improve the prediction accuracy. Compared with all-pairs correlations in RAFT-Stereo\cite{raft-stereo}, the GEV can provide non-local information and scene prior knowledge, thus the prediction error decreases evidently. RAFT-Stereo uses a starting disparity initialized to zero, thus increasing the number of iterations to reach optimal results. In contrast, we apply the $soft \; argmin$ to regress an initial starting disparity from GEV, which speeds up the convergence and slightly reduces the prediction error. To further explicitly constrain GEV during training, we use ground truth disparity to supervise GEV, deriving accurate GEV and starting disparity. When processed by the 3D regularization network, the GEV suffers from the over-smoothing problem at boundaries and tiny details. To complement local correlations, we combine the GEV and all-pairs correlations to form a combined geometry encoding volume (CGEV). The proposed CGEV, denoted as IGEV-Stereo, achieves the best performance.

\textbf{Number of Iterations.} Our IGEV-Stereo achieves excellent performance even when the number of iterations is reduced. As shown in Tab.~\ref{tab:iter}, we report the EPE of our models and RAFT-Stereo on Scene Flow. Compared with all-pairs correlations in RAFT-Stereo, our GEV can provide more accurate geometry and context information. Thus when the number of iterations is reduced to 1, 2, 3 or 4, our IGEV-Stereo (G) can outperform RAFT-Stereo with the same number of iterations by a large margin, such as surpassing RAFT-Stereo by 54.63\% at 1 iteration. When regressing an initial disparity $\mathbf{d}_0$ from GEV and supervising it, we obtain an accurate initial disparity to update and thus the prediction error can decrease evidently. Finally, when changing the number of iterations, our full model, denoted as IGEV-Stereo, achieves the best performance, which surpasses RAFT-Stereo by 69.44\% at 1 iteration. From Tab.~\ref{tab:iter}, we can observe that our IGEV-Stereo achieves the state-of-the-art performance even with few iterations, enabling the users to trade off time efficiency and performance according to their needs.

\textbf{Configuration Exploration.} Tab.~\ref{tab:ablation} shows results of different configurations.
Even constructing a 1/8 resolution GEV that takes only 5ms extra, our method still achieves state-of-the-art performance with an EPE of 0.49 on Scene Flow. When using the backbone with more parameters, i.e. MobileNetV2 120d and ConvNeXt-B\cite{convnext}, the performance can be improved.

\subsection{Comparisons with State-of-the-art}
We compare IGEV-Stereo with the published state-of-the-art methods on Scene Flow, KITTI 2012 and 2015. On Scene Flow test set, we achieve a new SOTA EPE of 0.47, which surpasses CSPN\cite{cspn} and LEAStereo\cite{leastereo} by 39.74\%. Compared to the classical PSMNet\cite{psmnet}, our IGEV-Stereo not only achieves $2\times$ better accuracy, but is also faster than it. Quantitative comparisons are shown in Tab.~\ref{tab:sceneflow}. We evaluate our IGEV-Stereo on the test set of KITTI 2012 and 2015, and the results are submitted to the online KITTI leaderboard. As shown in Tab.~\ref{tab:kitti}, we achieve the best performance among the published methods for almost all metrics on KITTI 2012 and 2015. At the time of writing, our IGEV-Stereo ranks $1^{st}$ on the KITTI 2015 leaderboard compared with over 280 methods. On KITTI 2012, our IGEV-Stereo outperforms LEAStereo\cite{leastereo} and RAFT-Stereo\cite{raft-stereo} by 10.00\% and 10.93\% on Out-Noc under 2 pixels error threshold, respectively. On KITTI 2015, our IGEV-Stereo surpasses CREStereo\cite{crestereo} and RAFT-Stereo\cite{raft-stereo} by 5.92\% and 12.64\% on D1-all metric, respectively. Specially, compared with other iterative methods such as CREStereo\cite{crestereo} and RAFT-Stereo\cite{raft-stereo}, our IGEV-Stereo not only outperforms them, but is also $2\times$ faster. Fig. \ref{fig:kitti} shows qualitative results on KITTI 2012 and 2015. Our IGEV-Stereo performs very well in reflective and detailed regions.

We evaluate the performance of IGEV-Stereo and RAFT-Stereo in the ill-posed regions, shown in Tab. \ref{tab:reflective}. RAFT-Stereo lacks non-local knowledge and thus has difficulties tackling local ambiguities in ill-posed regions. Our IGEV-Stereo can well overcome these problems. IGEV-Stereo ranks $1^{st}$ on KITTI 2012 leaderboard for reflective regions, which outperforms RAFT-Stereo by a large margin. Specially, our method performs better using only 8 iterations than RAFT-Stereo using 32 iterations in the reflective regions.

\begin{figure*}
\centering
\includegraphics[width=0.9\textwidth]{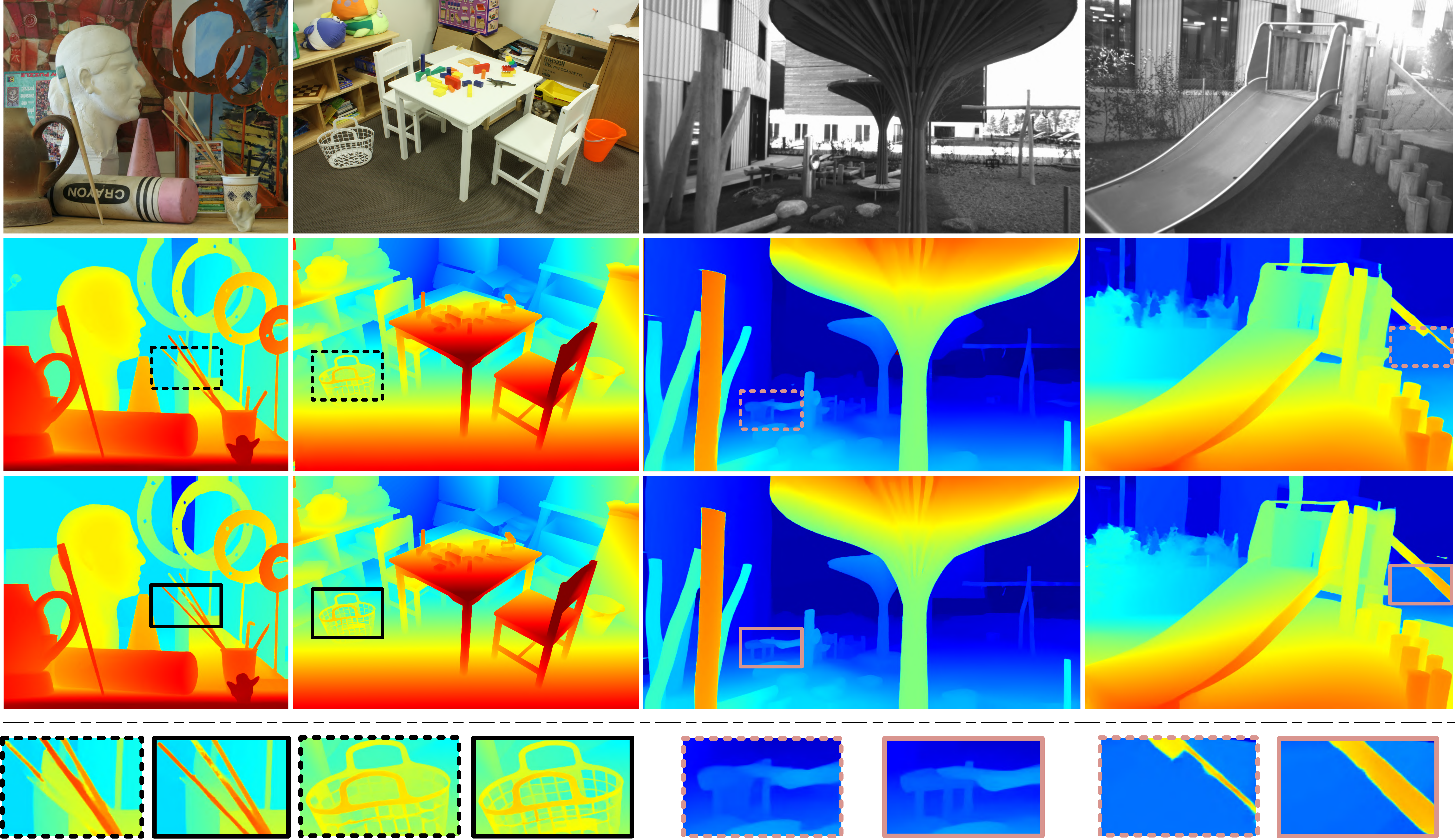} 
\caption{Generalization results on Middlebury 2014 and ETH3D. The second and the third rows are the results of RAFT-Stereo\cite{raft-stereo} and our IGEV-Stereo, respectively. Our IGEV-Stereo exhibits better details for fine-structured objects.}
\label{fig:middle_eth}
\vspace{-5pt}
\end{figure*}

\begin{figure}
\centering
\includegraphics[width=1.0\linewidth]{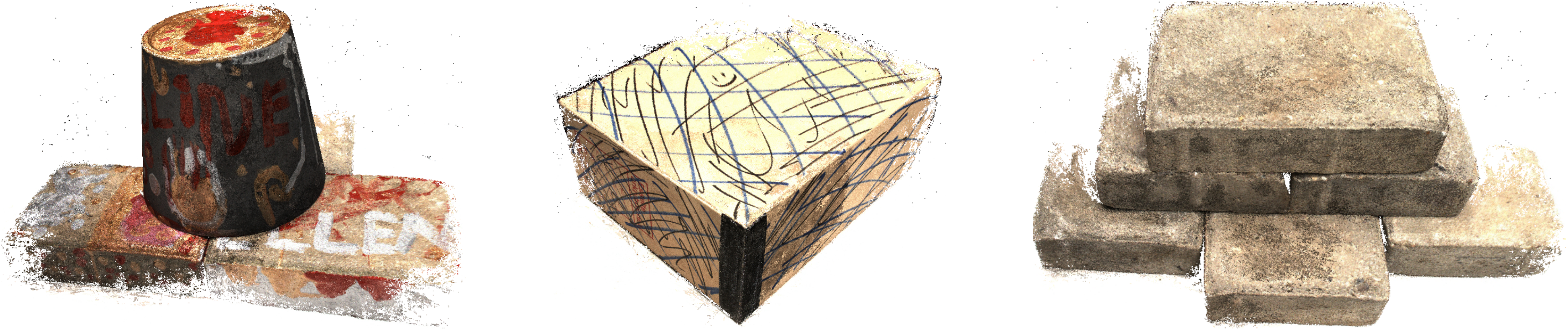} 
\caption{Visualization of results on DTU \cite{dtu} test set.}
\label{fig:dtu}
\vspace{-10pt}
\end{figure}

\begin{table}
  \centering
  \begin{tabular}{lccc}
    \toprule
    \multirow{2}{*}{Model} &\multicolumn{2}{c}{Middlebury} &\multirow{2}{*}{ETH3D}\\
    & half & quarter & \\
    \midrule
    CostFilter\cite{costfilter} &40.5 &17.6 &31.1\\
    PatchMatch\cite{patch} &38.6 &16.1 &24.1\\
    SGM\cite{sgm} &25.2 &10.7 &12.9\\
    \midrule
    PSMNet\cite{psmnet} &15.8 &9.8 &10.2 \\
    GANet\cite{ganet} &13.5 &8.5 &6.5 \\
    DSMNet\cite{dsmnet} &13.8 &8.1 &6.2 \\
    STTR\cite{sttr} &15.5 &9.7 &17.2 \\
    CFNet\cite{cfnet} &15.3 &9.8 &5.8\\
    FC-GANet\cite{fc} & 10.2 &7.8 &5.8 \\
    Graft-GANet\cite{graftnet} & 9.8 &- &6.2 \\
    RAFT-Stereo\cite{raft-stereo} & {8.7} &{7.3} &\cellcolor{yellow!25}\textbf{3.2} \\
    IGEV-Stereo (Ours) & \cellcolor{yellow!25}\textbf{7.1} &\cellcolor{yellow!25}\textbf{6.2} &{3.6} \\
    \bottomrule
  \end{tabular}
  \vspace{-2mm}
  \caption{
   Synthetic to real generalization experiments. All models are trained on Scene Flow.
   The 2-pixel error rate is used for Middlebury 2014, and 1-pixel error rate for ETH3D.}
  \label{tab:generalization}
  \vspace{-10pt}
\end{table}

\subsection{Zero-shot Generalization}
Since large real-world datasets for training are difficult to obtain, the generalization ability of stereo models is crucial. We evaluate the generalization performance of IGEV-Stereo from synthetic datasets to unseen real-world scenes. In this evaluation, we train our IGEV-Stereo on Scene Flow using data augmentation, and directly test it on the Middlebury 2014 and ETH3D training sets. As shown in Tab.~\ref{tab:generalization}, our IGEV-Stereo achieves state-of-the-art performance in the same zero-shot setting. Fig.~\ref{fig:middle_eth} shows a visual comparison with RAFT-Stereo, our method is more robust for textureless and detailed regions.

\subsection{Extension to MVS}
We extend our IGEV to multi-view stereo (MVS), i.e. IGEV-MVS. We evaluate our IGEV-MVS on the DTU\cite{dtu} benchmark. DTU is an indoor multi-view stereo dataset with 124 different scenes and 7 different lighting conditions. Following the IterMVS\cite{itermvs}, the DTU is split into training, validation and testing set. We use an image resolution of $640\times512$ and the number of input images is N=5 for training. We train IGEV-MVS on DTU for 32 epochs. For evaluation, image size, number of views and number of iterations are set to $1600\times1152$, 5 and 32 respectively. Compared with IGEV-Stereo, IGEV-MVS removes context network that means that ConvGRUs does not access context stream. As shown in Tab.~\ref{tab:dtu}, our IGEV-MVS achieves the best overall score, which is an average of completeness and accuracy. Especially, compared with PatchmatchNet\cite{patchmatchnet} and IterMVS\cite{itermvs}, our IGEV-MVS achieves 8.0\% and 10.7\% relative improvements on the overall quality.

\begin{table} \footnotesize
  \centering
  \begin{tabular}{lccc}
    \toprule
    Method &Acc.(mm)$\downarrow$ &Comp.(mm)$\downarrow$ &Overall(mm)$\downarrow$ \\
    \midrule
    Camp\cite{camp} &0.835 &0.554 &0.695 \\
    Furu\cite{furu} &0.613 &0.941 &0.777 \\
    Tola\cite{tola} &0.342 &1.190 &0.766 \\
    Gipuma\cite{gipuma} &\cellcolor{yellow!25}\textbf{0.283} &0.873 &0.578 \\
     \midrule
     MVSNet\cite{mvsnet} & 0.396 & 0.527 &0.462 \\
     R-MVSNet\cite{r-mvs} & 0.383 & 0.452 &0.417 \\
     P-MVSNet\cite{p-mvs} & 0.406 & 0.434 &0.420 \\
     Point-MVSNet\cite{point-mvs} & 0.342 & 0.411 &0.376 \\
     Fast-MVSNet\cite{fast-mvs} & 0.336 & 0.403 &0.370 \\
     AA-RMVSNet\cite{aa-rmvs} & 0.376 & 0.339 &0.357 \\
     CasMVSNet\cite{cascade} & 0.325 & 0.385 &0.355 \\
     UCS-Net\cite{ucsnet} & 0.338 & 0.349 &0.344 \\
     CVP-MVSNet\cite{cvp-mvs} & 0.296 & 0.406 &0.351 \\
     MVS2D\cite{mvs2d} & 0.394 & 0.290 &0.342 \\
     PatchmatchNet\cite{patchmatchnet} & 0.427 & \cellcolor{yellow!25}\textbf{0.277} &0.352 \\
     IterMVS\cite{itermvs} & 0.373 & 0.354 &0.363 \\
     CER-MVS\cite{cer-mvs} &0.359 & 0.305 &0.332\\
     IGEV-MVS (Ours) &0.331 &0.316 &\cellcolor{yellow!25}\textbf{0.324}\\
     
    \bottomrule
  \end{tabular}
  \vspace{-2mm}
  \caption{
  Quantitative evaluation on DTU. Methods are separated into two categories (from top to bottom): traditional and trained on DTU.}
  \label{tab:dtu}
  \vspace{-15pt}
\end{table}

\section{Conclusion and Future Work}
We propose Iterative Geometry Encoding Volume (IGEV), a new deep network architecture for stereo matching and multi-view stereo (MVS). The IGEV builds a combined geometry encoding volume that encodes geometry and context information as well as local matching details, and iteratively indexes it to update the disparity map. Our IGEV-Stereo ranks $1^{st}$ on KITTI 2015 leaderboard among all the published methods and achieves state-of-the-art cross-dataset generalization ability. Our IGEV-MVS also achieves competitive performance on DTU benchmark. 

We use a lightweight 3D CNN to filter the cost volume and obtain GEV. However, when dealing with high-resolution images that exhibit a large disparity range, using a 3D CNN to process the resulting large-size cost volume can still lead to high computational and memory costs. Future work includes designing a more lightweight regularization network. In addition, we will also explore the utilization of cascaded cost volumes to make our method applicable to high-resolution images.

\textbf{Acknowledgement.} We thank all the reviewers for their valuable comments. This research is supported by National Natural Science Foundation of China (62122029, 62061160490) and Applied Fundamental Research of Wuhan (2020010601012167).

{\small
\bibliographystyle{ieee_fullname}
\bibliography{egbib}
}

\end{document}